\documentclass[preprint]{revtex4-1}
\usepackage{blindtext}
\usepackage{subfiles}

\usepackage{graphicx}
\usepackage{dcolumn}
\usepackage{bm}

\usepackage{amssymb}
\usepackage{epstopdf}
\usepackage{color}
\usepackage{soul}
\usepackage{xr}
\usepackage{longtable}
\usepackage{subfigure}
\usepackage{amsmath}
\usepackage{appendix}
\newcommand{\bs}{\bm{s}}

\newcommand{\bt}{\bm{t}}
\newcommand{\bJ}{\bm{J}}

\newcommand{\bvartheta}{\bm{\vartheta}}
\newcommand{\btheta}{\bm{\theta}}

\newcommand{\hbs}{\hat{\bm{s}}}

\newcommand{\hbtheta}{\hat{\bm{\theta}}}
\newcommand{\red}[1]{\textcolor{black}{#1}}

\newcommand{\balpha}{\bm{\alpha}}

\begin{document}


\title{Lost in Retraining: Roaming the Parameter Space of Exponential Families Under Closed-Loop Learning}

\author{Fariba Jangjoo}
 \affiliation{Kavli Institute for Systems Neuroscience, NTNU, Trondheim, Norway}
\author{Matteo Marsili}%
\affiliation{%
ICTP- International Centre for Theoretical Physics, Trieste, Italy\\
}%
\author{Yasser Roudi}%
\affiliation{%
Department of Mathematics, King's College London, London, UK
}%

\date{\today}

\begin{abstract}
Closed-loop learning is the process of repeatedly estimating a model from data generated from the model itself. It is receiving great attention due to the possibility that large neural network models may, in the future, be primarily trained with data generated by artificial neural networks themselves. We study this process for models that belong to exponential families, deriving equations of motions that govern the dynamics of \red{the} parameters. We show that \red{maximum likelihood estimation of the parameters} endows sufficient statistics with the martingale property and \red{that as a result} the process converges to absorbing states that amplify initial biases present in the data. However, we show that this outcome may be prevented if the data contains at least one data point generated from a ground truth model, 
by relying on maximum {\em a posteriori} estimation or by introducing regularisation. 
\end{abstract}

\pacs{Valid PACS appear here}
\maketitle

\textit{\textbf{Introduction}}--. 
\red{This paper focuses on the behaviour of statistical models and learning machines subject to {\em closed-loop} learning: the process in which models are repeatedly trained on data mainly generated by models themselves. When does this process converge and if so to what?} These questions have raised considerable recent interest in the context of 
Large Language models (LLM) and other state of the art neural networks~\cite{alemohammad2024self,bertrand2024stability,marchi2024heat,shumailov2024ai,guo2023curious,briesch2024largelanguagemodelssuffer,dohmatob2024modelcollapsedemystifiedcase,vu2025happens}. The success of LLMs relies on the availability of huge amounts of high-quality human-generated data. Yet the supply of such data has been argued to be running out~\cite{villalobos2024will}. At the same time, more and more AI-generated data is populating the Internet, thereby polluting the datasets with which models will be trained. It has been argued that this process of ``data-autophagy'' is pathologic with a severity that depends on the fraction of synthetic data used in training~\cite{alemohammad2024self,bertrand2024stability,briesch2024largelanguagemodelssuffer}. A possible outcome of this is {\em model collapse}~\cite{shumailov2024ai} -- a phenomenon observed also in Generative Adversarial Networks~\cite{thanh2020catastrophic}, diffusion models~\cite{biroli2024dynamical} and Variational Autoencoders~\cite{shumailov2024ai} and LLM~\cite{guo2023curious} -- whereby the model generates only a limited variety of outputs, failing to capture the full diversity of the training data. Marchi~{\em et al.}~\cite{marchi2024heat} find that closed-loop learning can also lead to the opposite outcome, where models generate random outputs. As we shall argue in the conclusions, close-loop learning is also of relevance for a wide variety of other phenomena, ranging from the evolution of cultures to rational herding~\cite{witzel2012origins,bikhchandani1992theory}.

Here, we study closed-loop learning within a simple yet general setting, that of Exponential Family (EF). 
Distributions in EF take the form (see e.g. \cite{bishop2023deep} sec. 3.4)
\begin{equation}
\label{eq:expfam}
f(s|\btheta)=\frac{1}{Z(\btheta)}e^{\sum_a\theta_a\phi_a(s)},~~~ Z(\btheta)=\sum_{s} e^{\sum_a\theta_a\phi_a(s)},
\end{equation}
\red{where the random variable $s$ takes value in $\mathcal{S}$, $\btheta\in\mathbb{R}^D$ is a vector of $D$ parameters with components $\theta_a$. Each $\phi_a(s)$, $a=1,\cdots,D$, is a bounded functions of $s$, 
and $Z$ is the partition function. We aim to understand how the parameters evolve if they are fitted to data generated by the model with the previous set of parameters, including potentially data generated by an external source.} As we show, this simplified setting offers a transparent and fully controllable rationalization of a set of phenomena that occur when retraining models on model-generated data, with the potential to unearth general mechanisms that apply to a broader domain. 

\red{We first study the dynamics of the model parameters follow in closed-loop learning when they are estimated using Maximum Likelihood (ML) - a widely used approach- and from data exclusively drawn from the model itself. In the case of ML, the relevant variables are the sufficient statistics of the model \cite{casella2024statistical} which, as we show, acquire a martingale property. This property in turn will be shown to lead to model collapse. We then discuss how this collapse may be avoided. To do this, we consider the more general case in which parameter learning may go beyond ML and may involve a prior distribution/regularization (as in MAP inference). We also consider the possibility that external data (e.g. from a ground truth distribution taken to be human generated data in LLMs) may also be supplied alongside self-generated data during repeated training. For this general setting, we derive the Fokker-Planck equations governing the stochastic dynamics of the model parameters during closed-loop learning. Finding the the stationary state distribution of the Fokker-Planck equations, we find that both the presence of priors/regulations and external data may prevent model collapse. In particular, we show that even adding one single datapoint from outside may be sufficient for avoiding model collapse. Surprisingly, these happen even in the limit where, at each retraining iteration, the amount of available self-generated data approaches infinity. This is surprising because, in standard (non-closed-loop) learning, the differences between MAP estimation and maximum likelihood, as well as the effects of priors and regularization, become negligible.}

\textit{\textbf{Closed-loop learning in EF}}--.
\red{Consider the following dynamics in the space of parameters $\btheta\in \mathbb{R}^d$. The dynamics starts at $\btheta_0$ which we take to be the result of fitting the model to some data. From the initial parameter vector,} at iteration $t\ge 0$,
\begin{description}
\item[1] draw $M$ independent samples, ${\hat s}^1(t),\ldots,{\hat s}^M(t)$, from Eq.~\eqref{eq:expfam} with $\btheta=\btheta_t$ and $m$ samples ${\hat s}^{M+1}(t),\ldots,{\hat s}^{M+m}(t)$ generated from  a distribution $q(s)$. Denoting these samples as $\hat{\bs}(t)\equiv({\hat s}^1(t),\ldots,{\hat s}^{M+m}(t))$, their likelihood is
\[
p(\hat\bs(t)|\btheta_t)=\prod_{i=1}^M f(\hat s^i(t)|\btheta_t)\prod_{j=1}^m q(\hat s^j(t)).
\]
\item[2] Given $\hat \bs(t)$ find the parameters 
\begin{equation}
\label{eq:orig2}
  \btheta_{t+1}\equiv \hat\btheta(\hat \bs(t)) ={\rm arg}\max_{\btheta} \left[p(\hat \bs(t)|\btheta)p_0^u(\btheta)\right].
\end{equation}
  \item[loop] Set $t\to t+1$ and repeat.
\end{description}
When $u=0$ the parameter are inferred by Maximum Likelihood (ML) whereas for $u=1$ inference relies on maximum {\em a-posteriori} (MAP) inference with a prior $p_0(\btheta)$. 
This includes regularization schemes, e.g., Lasso~\cite{tibshirani1996regression}, that can be represented by an appropriate prior. With $m=0$ the dynamics is one in which data are generated uniquely from the model itself whereas with $m>0$ new data generated from a distribution $q(\bs)$ are also included. The distribution $q$ could also be the model generating the original data that determined $\btheta_0$ and reflect a {\it ground truth}

\red{The steps above yield a Markov dynamics on the parameters: at each step the new data is generated consistently with the data just learned. In what follows, we first study this dynamics for $u=m=0$, that is ML and completely self-generated data. We then study the general case.}

\textit{\textbf{Closed-loop learning under ML ($m=u=0$)}}--. The components $\phi_a(s)$ in Eq.~\eqref{eq:expfam} are the sufficient statistics of the model. This means that for $m=u=0$ given a sample $\hat \bs$, the ML estimates, $\hat\btheta(\hat \bs)$, depend solely on the empirical averages 
\begin{equation}
\bar\phi_a(\hat \bs)=\frac{1}{M}\sum_{l=1}^M \phi_a(\hat s^l)\,.
\end{equation}
and are the solutions of the following set of equations
\begin{equation}
\label{eq:Ephi}
\langle \bar\phi_a(\bs) \rangle_{\hat\btheta}=\bar\phi_a(\hat \bs),\qquad a=1,\ldots,d\,.
\end{equation}
where $\bs = (s^1,\dots,s^M)$, for $s^{l} \in \mathcal{S}$ and $\langle \ldots\rangle_{\btheta}$ 
is expectation with respect to $p(\bs|\btheta)=\prod_{l=1}^M f(s^l|\btheta)$.
 
In the closed-loop dynamical process described above, $\hat\bs(t)$ determines $\bar\phi_a(\hat \bs(t))$ and thus, via Eq.~\eqref{eq:Ephi}, determines $\btheta_t=\hat\btheta(\hat \bs(t))$ and the distribution $p(\cdot|\btheta_t)$. Consequently, given $\hat\bs(t)$ and defining $\bar\phi_{a,t}\equiv \bar\phi_a(\hat \bs(t))$, we can write
\begin{equation}
\label{eq:martingale}
\langle\bar\phi_{a,t+1}|\bar\phi_{a,t}\rangle_{\btheta_t}
=\bar\phi_{a,t}\,.
\end{equation}
where $\langle\ldots|\bar\phi_{a,t}\rangle_{\btheta_t}$ is the average over all possible samples $\hbs(t+1)$ generated at fixed $\btheta_t$, which as described above is determined by $\bar\phi_{a,t}$. 

\red{By definition, Eq.~(\ref{eq:martingale}) means} that $\bar\phi_{a,t}$ is a martingale~\cite{roldan2023martingales}. 
\red{As shown in Feller~\cite{feller1968introduction} (see also Appendix~\ref{appA}), a Markov chain $\bar\phi_{a,t}$ that is also a martingale in a bounded state space converges to absorbing states.} 
In other words, the fluctuations of $\bar\phi_{a,t}$ vanish, \red{i.e. $\phi_a(s)$ attains the same value for all data points $\hat\bs(t)$, and}
\begin{equation}
\lim_{t\to\infty} f(s|\btheta_t)=c\prod_a\delta(\phi_a(s)-\bar\phi_{a,\infty})
\end{equation}
with $c$ a normalizing constant. \red{This is what is meant by model collapse: while the parameters may be initialized at any value, closed-loop learning prevents the model to represent anything but few possibilities. Below we illustrate this for three example models.}

Consider an \red{Ising model defined over spin configurations} $s=(\sigma_1,\ldots,\sigma_N)$ with $\sigma_i=\pm 1$ as
\begin{equation}
\label{eq:FCIM}
f(s|\theta_1,\theta_2)=\frac 1 Z \exp\Big[\theta_1 \sum_i \sigma_i +\frac{\theta_2}{N}\sum_{i<j} \sigma_i \sigma_j\Big]\,.
\end{equation}
\begin{figure}
    \centering
    \includegraphics[width=0.85\linewidth]{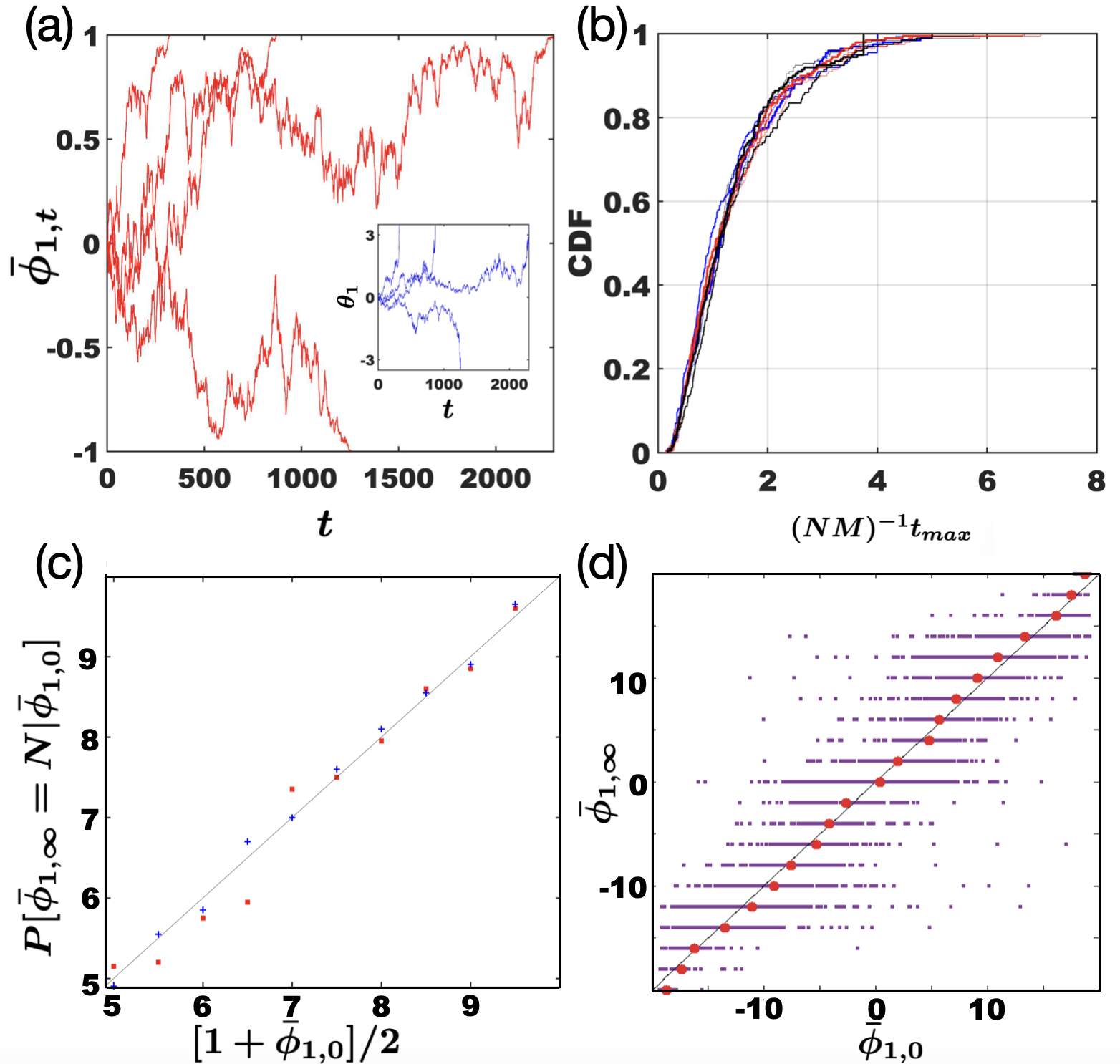}
    \caption{Closed-loop dynamics and model collapse for $m=u=0$ in the Ising model (Eq.~\eqref{eq:FCIM}). (a), (b) and (c) refer to 
    the case $\theta_2=0$ whereas in (d), $\theta_2\neq 0$. (a) Example trajectories of $\phi_{1,t}$, starting at $\phi_{1,0}=0$; $N=20$ and $M=50$. Corresponding trajectories for $\theta_1$ shown in the inset. (b) Cumulative distribution for the convergence time to fully polarized states normalized by $NM$ for $N=5,10,20$ (blue, red,black lines), and $M=50,100,200$ (thinnest to thickest lines). (c) Probability of convergence to all spins up, as a function of $(1+\bar{\phi}_{1,0})/2$ (see Eq.~\eqref{eq:PphiisingJ0}); each point is calculated from $200$ simulations with the same initial $\bar{\phi}_{1,0}$, with $N=10$; $M=50$ (red squares) and $M=100$ (blue circles). (d) The final values of $\bar{\phi}_1$ vs. initial values for $\theta_2\neq 0$, for $100$ different simulations with $N=20$ and $M=100$. $\theta_2$ is set to zero at  $t=0$ but updated afterwards. Full circles are averages of $\bar\phi_{1,0}$ for the same value of $\bar\phi_{1,\infty}$.}
    \label{fig:Fig1}
\end{figure}
With $\theta_2=0$ (no interaction) 
only  $\phi_{1}(s) = \sum_i \sigma_i$ is relevant and $\bar \phi_1(\hat \bs)$ 
converges to $\pm N$, with (see Appendix~\ref{appA})
\begin{equation}
\label{eq:PphiisingJ0}
\lim_{t\to\infty}P\{\bar \phi_{1,t}= N |\bar \phi_{1,0}\} = \frac{\bar \phi_{1,0}+N}{2N}\,;
\end{equation}
Correspondingly, $\theta_{1,t}$ diverges to $\pm\infty$ and the distribution $f(s|\theta_1,\theta_2=0)$ converges to a fully polarized state where in the sample $\hat\bs$, either $\sigma_i^l=+1$ or $\sigma_i^l=-1$ for all $l$. Fig.~\ref{fig:Fig1}(a) shows examples of this dynamics. The empirical cumulative distributions of $t_{max}$, the time it takes for the dynamics to converge, shown in Fig. \ref{fig:Fig1}(b), suggests that $t_{max}\sim NM$. Fig.~\ref{fig:Fig1}(c) shows simulations results confirming Eq.~(\ref{eq:PphiisingJ0}).

If instead $\theta_2 \neq 0$, \red{any $\bar\phi_{1,\infty} =N-2k$ with $k=0,1,2,\cdots N$}, can be an absorbing state. For $|\bar\phi_{1,\infty}|<N$, this occurs because both $\theta_1$ and $\theta_2$ diverge, i.e. $\theta_2\to-\infty$ with a finite ratio $\theta_1/\theta_2=-\bar\phi_{1,\infty}/N$ \red{see SI}. In this case, $f(s|\theta_1,\theta_2)$ converges to a uniform distribution on \red{all configurations $s$ with the same value of $\phi_{1}(s)$.} 
\red{Fig.~\ref{fig:Fig1}(d) shows that for} $\theta_{2,t=0}=0$, the limiting value of $\bar\phi_{1,\infty}$ is not far from the initial value $\bar\phi_{1,0}$.

To further illustrate \red{model collapse in closed-loop learning, we next consider a spin model where $\theta_s$} is an independent parameter for each of the $d=2^n$ spin configurations $s=(\sigma_1,\ldots,\sigma_N)$. The corresponding operator in Eq.~\eqref{eq:expfam} is thus $\phi_s(s')=\delta_{s,s'}$.
When $\theta_s$ are drawn from a Gaussian distribution, this model coincides with the {\em Random Energy Model}~\cite{derrida1981random}. \red{Yet here we focus on the closed-loop learning setting with $m=u=0$, where $\btheta_{t+1}$ are the ML estimates inferred from a $M$ samples $\hat\bs(t)$ drawn i.i.d. from the model with parameters $\btheta_t$.} For $M\ll 2^N$ this is an over-parametrized model.
The average of $\phi_s$ over the data $\hat\bs(t)$ is $\bar\phi_{s,t}=k_{s,t}/M$, where $k_{s,t}$ is the number of times configuration $s$ occurs in the data $\hat\bs(t)$, and $f(s|\hat\btheta(\hat\bs (t))=\frac{k_{s,t}}{M}$. \red{A state $s$ which is not observed at time $t$, i.e. if $k_{s,t}=0$, will thus never be observed forever after. The number of states with $k_{s,t}>0$ is thus a non-increasing function of time.} The asymptotic state will be one where $k_{s^*,t}=M$ for one state and $k_{s,t}=0$ for all $s\neq s^*$, which correspond to the two absorbing states $\bar \phi^{+}_s = M$ and $\bar \phi^{+}_s = 0$. Furthermore, it can be shown that (see Appendix~\ref{appA})
\begin{equation}
\label{eq:rem}
\lim_{t\to\infty} P\{k_{s,t}=M| k_{s,0}\}=\frac{k_{s,0}}{M}\,.
\end{equation}
Summarizing, \red{closed-loop learning in EF based on ML and not data from outside ($m=0$) leads to a reduction of the diversity of the data. How severe this reduction is depends on the model and the initial conditions.}

\textit{\textbf{The general case and continuous time dynamics}}--.
Let us now consider the general case with an arbitrary integer value of $m$ and $u\in\{0,1\}$.

Between two consecutive iterations, we expect that the difference $\btheta_{t+1}-\btheta_t$ to be small, i.e. of the same order of the error on the ML estimator $\delta\hat\btheta\propto M^{-1/2}$. So for large $M$, we expect the evolution of $\btheta_t$ to be well approximated by a continuous time dynamics. We thus consider the transition matrix between $t$ and $t+1$
\begin{equation}
p(\btheta_{t+1}\!=\!\btheta'|\btheta_t\!=\!\btheta)=
\sum_{\hat \bs}p(\hat \bs|\btheta)\delta\left(\btheta'-\btheta_{t+1}(\hat \bs)\right)
\end{equation}
where $\btheta_{t+1}(\hat \bs)$ is given by Eq.~(\ref{eq:orig2}) and the probability of $\hat\bs$ may depend both on $\btheta=\btheta_t$ and on $q$, as detailed in \red{step$\mathbf{1}$ of the dynamics described above}. In the limit $M\to\infty$ the transition matrix is dominated by a sharp maximum close to the ML estimator $\hat\btheta$. \red{Expanding around this point (see Appendix~\ref{appB}), to the leading order in $M$, $\btheta_{t+1}$ is well approximated by a Gaussian variable with covariance $J^{-1}(\btheta_t)/M$, where}
\begin{equation}
\label{FIM}
J_{a,b}(\btheta)=\frac{\partial^2}{\partial\theta_a\partial\theta_b}\log Z=\langle \phi_a\phi_b\rangle_{\btheta}- \langle\phi_a\rangle_{\btheta}\langle\phi_b\rangle_{\btheta}
\end{equation}
\red{is the Fisher Information Matrix (FIM). The Gaussian has a mean $\btheta_t+A(\btheta_t)/M$ where}
\begin{eqnarray}
\label{eq:drift}
&A_a(\btheta) = -\sum_b J^{-1}_{a,b}\frac{\partial U}{\partial\theta_b}\\
&U(\btheta) = \frac 1 2 \log{\rm det}J-u\log p_0(\btheta)+m\,{\rm D_{KL}}(q|\btheta)\label{eq:U}.
\end{eqnarray}
\red{Here ${\rm D_{KL}}(q|\btheta)=\sum_sq(s)\log\frac{q(s)}{f(s|\btheta)}$ is} the Kullback-Leibler divergence between $q(s)$ and $f(s|\btheta)$. Note that in the ML case ($u=m=0$) the drift term (Eq \ref{eq:drift}) is non-zero, as a consequence of the fact that ML estimators are, in general, biased ~\cite{cox1979theoretical}.

In the rescaled time $\tau=t/M$, the vector of parameters satisfies the (Ito) stochastic differential equation~\cite{gardiner2009stochastic} (see Appendix~\ref{appB})
\begin{eqnarray}
d\btheta_\tau & = & A(\btheta_\tau)d\tau+dW_\tau \label{eq:dtheta}\\
\langle dW_\tau dW_{\tau'}\rangle & = &  J^{-1}(\btheta_\tau)\delta(\tau-\tau')d\tau\,. \label{eq:dW}
\end{eqnarray}
\red{This dynamics is fully consistent with the model collapse discussed in the previous section for $m=u=0$. Indeed Eqs.~\eqref{eq:dtheta}-\eqref{eq:dW} can be used to show that the evolution of the sufficient statistics $\varphi_a\equiv\langle\phi_a(\bs)\rangle_{\btheta}$
is a purely diffusive dynamics (see Appendix~\ref{appB}), eventually hitting and getting absorbed in the boundaries.} 

The continuous time dynamics indeed shows that closed-loop learning leads to a contraction of the configuration space. For example, for an Ising model with $\theta_2=0$, the dynamics of the magnetization $\mu\!=\!\langle\bar\phi_1\rangle_{\btheta}/N\!=\!\langle\sum_i\sigma_i\rangle_{\btheta}/N$ per spin is given by $d\mu=\sqrt{1-\mu^2}dW$, where $dW$ is the differential of a Wiener process. This dynamics clearly has two absorbing states at $\mu=\pm 1$. Indeed $\langle \mu^2\rangle_{\theta_\tau}=1-(1-\langle \mu^2\rangle_{\theta_0})e^{-\tau}$. More generally, \red{one can  consider the evolution of the entropy
\begin{equation}
    S(\bm{\varphi})=\log Z-\sum_a\theta_a\varphi_a\,,
\end{equation}
as a function of the conjugate variables $\bm{\varphi}$ to show that
the configuration space, indeed, shrinks. To see this, note that $-S(\bm{\varphi})$ is the Legendre transform of $\log Z$, in the conjugate coordinates $\bm{\varphi}$, and hence $\frac{\partial S}{\partial\varphi_a}=-\theta_a$ and $\frac{\partial^2 S}{\partial\varphi_b\partial\varphi_c}=-J_{b,c}^{-1}$~\cite{amari2016information}. The differential of the entropy is thus $dS=-\sum_a\theta_ad\varphi_a-D\,d\tau$. Consequently, the expected value of the entropy changes with time $\tau$ as
\begin{equation}
\label{eq:entred}
    \langle S\rangle_{\bm{\varphi}_\tau}=\langle S\rangle_{\bm{\varphi}_0}-\tau\, D
\end{equation}
diverges to $-\infty$ as $\tau\to\infty$. Importantly, Eq.~\eqref{eq:entred} shows that the entropy reduces at a faster rate for higher dimensional models.}

\begin{figure}
    \centering
    \includegraphics[width=0.95\linewidth]{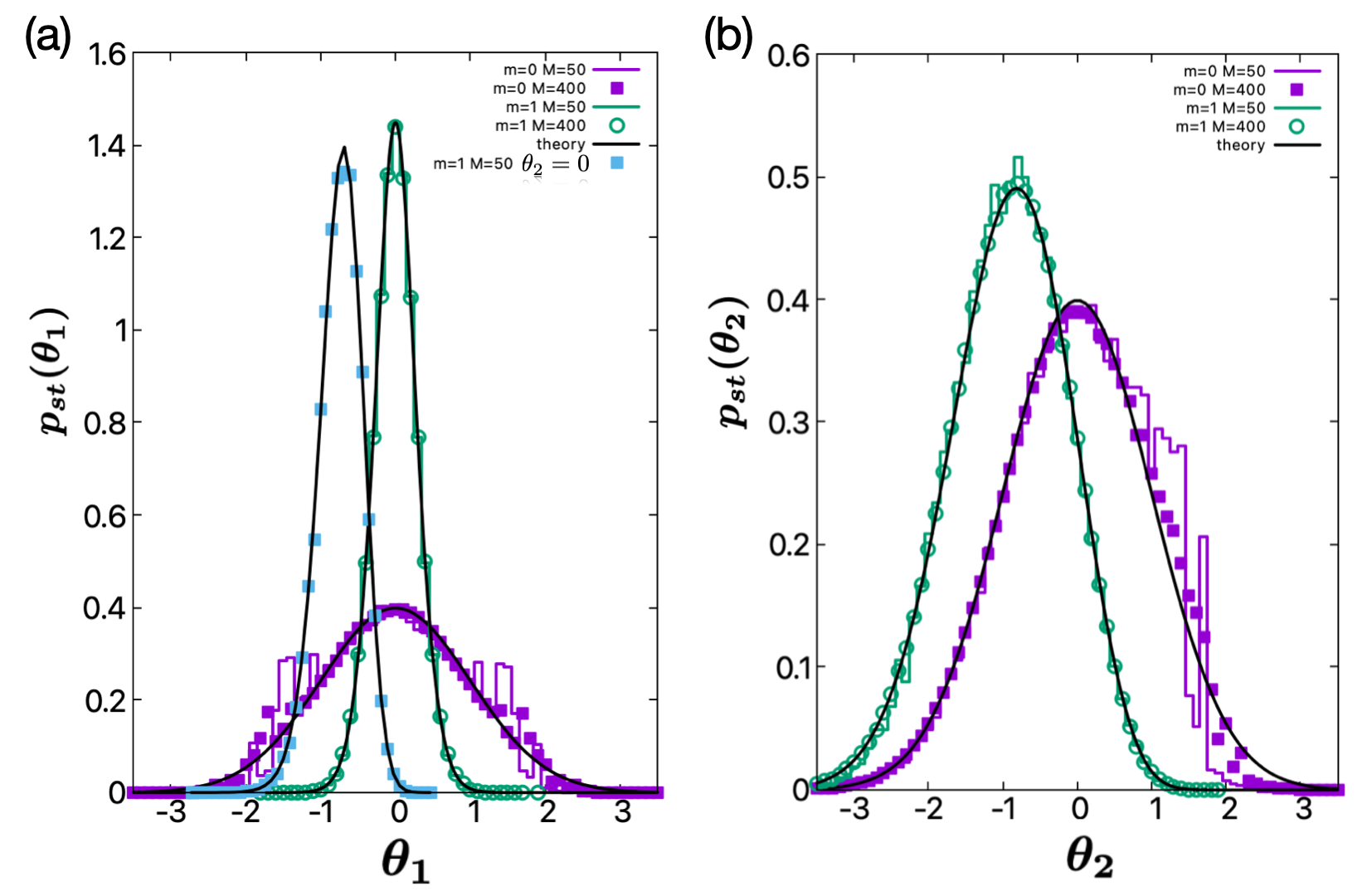}
    \caption{\red{Numerical simulation confirming Eq.~\eqref{eq:pstat} and prevention of model collapse for the Ising model. (a) The distribution of the field $\theta_1$.(b) The distribution of the coupling $\theta_2$. We consider three cases all with $N=10$: {\em i}) $\theta_2=0$ and $M=50$, with one data point with $\sum_i \sigma_i = -6$ being added ($m=1$) while $u=0$ (ML); {\em ii}) $\theta_2\neq 0$, $m=0$ and $M=50$ and $400$, with Gaussian priors on both $\theta_1$ and $\theta_2$ with zero mean and variance $2$ (which is equivalent to $L2$ regularization with parameter $1/4$); {\em iii}) $\theta_2\neq 0$, $M=50$ and $400$, with the same priors and the addition of one data point ($m=1$) with $\sum_i \sigma_i = 0$. In all cases model collapse is avoided as predicted by Eq.~(\ref{eq:pstat}) with a good agreement 
    (see full black lines). Data is obtained from simulations of $10^7$ iterations of the dynamics.}}    
    \label{fig:test_Ising}
\end{figure}

\red{In the presence of priors or samples drawn from $q(s)$ ($u=1$ or $m>0$), the asymptotic behavior of the dynamics can be studied by deriving the 
Fokker-Planck equation associated to Eq.~\eqref{eq:dtheta} using standards techniques (see Appendix~\ref{appB}). This shows that the dynamics converges to a stationary state}
\begin{equation}
\label{eq:pstat}
p_{st}(\btheta) = c\, p_0^{2u}(\btheta)e^{-2m{\rm D_{KL}}(q|\btheta)}.
\end{equation}
\red{assuming the normalization constant $c$ is finite. Model collapse can be identified when it is not. In particular, consistent with the previous section, this happens when $u=m=0$. However, even the addition of one external data point ($m=1$) or turning on the prior ($u=1$) may change this. For example, for the Ising model in Eq.~\eqref{eq:FCIM} with $\theta_2=0$ under ML ($u=0$) and with $m=1$, model collapse will be avoided, as confirmed numerically in Fig.~\ref{fig:test_Ising}a. In Fig.~\ref{fig:test_Ising}, we also numerically confirm the analytical expression in Eq.~\eqref{eq:pstat} for MAP ($u\neq 0$) for the Ising model when $\theta_2\neq 0$, $m=0$ and $m=1$, showing that model collapse is again avoided.}

\red{In general, whether, or to what degree,} using MAP ($u=1$) or adding external data ($m>0$) helps avoiding model collapse depends on the details of the model, the choice of the prior and the distribution $q$. \red{To further demonstrate this, it is instructive to consider another example besides the Ising model discussed above. In Appendix~\ref{appC} we consider a Poisson distribution with mean $\vartheta$. In terms of $\theta=\log\vartheta$, the model takes the form in Eq.~\eqref{eq:expfam}}. Assuming an exponential prior $p_0(\vartheta)=\lambda e^{-\lambda\vartheta}$ and $q(s)$ a Poisson distribution with mean $\vartheta_0$, one can analytically derive the stochastic differential equation that $\vartheta$ follows, which turns out to be a Cox–Ingersoll–Ross process \cite{cox1985theory}. \red{With $m=0$, the stationary distribution does not exist and the system converges to the absorbing state $\vartheta_\infty=0$ even with $\lambda>0$. This can, however, be prevented even with a small amount of data from outside ($m>0$) for which the dynamics converges to a stationary state $p_{st}(\vartheta)$ which is a Gamma distribution with shape parameter $2m\vartheta_0$ which coincides with
Eq.~ \eqref{eq:pstat} for $\theta$. These analytical results for the Poisson model are compared and confirmed with simulations in Appendix Fig.~\ref{fig:m}, where we again show simulation examples where adding only one external data point prevents model collapse, even in the absence of priors or regularization.}


\textit{\textbf{Discussion}}--. \red{We studied closed-loop learning for a models in the exponential families, analytically deriving the equations governing the stochastic dynamics of the parameters
We found that in this setting,  as a result of the the martingale property in Eq. \eqref{eq:martingale}, model collapse occurs when inference is based on the maximization of the likelihood on data is entirely generated by the model itself. Remarkably, we showed that the introduction of even one single data point generated from a ground truth distribution, or adopting MAP estimates or introducing regularisation can avoid model collapse.} In classical statistical theory, the difference between MAP ($u=1$) and ML ($u=0$) estimates vanishes for $M\to\infty$. Likewise, polluting a sample with a finite number $m$ of data points drawn from a different distribution has no effect on the asymptotic behavior of statistical estimates. Our results, however, show that these conclusions do not hold in closed-loop learning. 

These results may help shed light in the current debate on closed-loop learning in large aritificial network models~\cite{alemohammad2024self,bertrand2024stability,marchi2024heat,shumailov2024ai,guo2023curious,briesch2024largelanguagemodelssuffer,dohmatob2024modelcollapsedemystifiedcase,vu2025happens}. \red{Although here the martingale property and its consequences were derived and studied for the exponential family, it may hold for other models as well, with similar consequences; see e.g. \cite{alemohammad2024self}. Yet,} it is important to acknowledge that these models are largely over-parametrized and operate in a regime that is very different from that of classical statistics~\cite{bartlett2021deep} assumed here. In spite of this, our analysis aligns with observations reported in more complex cases of much recent interest in AI~\cite{alemohammad2024self,bertrand2024stability,marchi2024heat,shumailov2024ai,guo2023curious,briesch2024largelanguagemodelssuffer,dohmatob2024modelcollapsedemystifiedcase,vu2025happens}. This suggests that our analysis can provide a starting point for addressing issues such as implicit biases arising from algorithms, architectures, and/or initial conditions in artificial neural networks \cite{rahaman2019spectral,francazi2023initial,jain2024bias}, providing intuition and testable prediction within an analytically controllable framework. In particular, our results suggest that while ML amplifies biases in the initial conditions, MAP or the introduction of regularization emphasizes biases implicit in algorithms and/or architectures. 
\red{Our analysis shows that FIM is what determines drift and diffusion in closed-loop learning. Studying the FIM in deep networks using techniques such as those of \cite{hayase2021spectrum, cao2020nonparametric} may thus help understanding the precise dynamics of closed-loop learning in such networks. We also note that deriving the continuous time dynamics here relied on the the fact that the FIM of EF is invertible. Generalizing our analysis to singular models where this does not hold~\cite{watanabe2013widely} is another interesting avenue of further research.}

\red{Dynamics similar to closed-loop learning is present in many domains including} word-of-mouth communication, repeated model based imitation \cite{baker2009action,najar2020actions,charpentier2020neuro} and rational herding~\cite{bikhchandani1992theory}, or cultural transmission across generations~\cite{witzel2012origins}.
Pre-literacy cultural evolution has been based on one generation learning from the previous one, leading to a corpus of ``data'' of, e.g., mythological nature that shares remarkable similarities across distant cultures~\cite{witzel2012origins}. The debate remains open as to whether these similarities stem from shared ancestry and historical migrations~\cite{witzel2012origins}, from innate archetypal structures or from universal cognitive mechanisms and biases~\cite{Boyer2001}. \red{In social contexts,} individuals learn from the observed actions of others, and their resulting behavior is observed by others. This mechanism potentially leads to ``informational cascades'' that have been argued to determine many collective phenomena~\cite{bikhchandani1992theory}. \red{In} prenatal development, the brain develops to a large extent in the absence of external stimuli, driven by its own spontaneous activity~\cite{martini2021spontaneous,wong1999retinal}. What is common in these phenomena is that learning takes place using data generated by another learnt model. Although the process discussed here is very \red{simple and stylized, it offers a platform for studying these phonomena} and for engineering models that implement specific functions or that feature pre-assigned structures of asymptotic states.

\textit{\textbf{Acknowledgments}}--. We are grateful for interesting discussions with Manfred Opper, Fernando Villegas and Claudio Arezzo. We also acknowledge the use of LLM at different stages of this research.

\bibliographystyle{h-physrev3}
\bibliography{ModeCollapse25}

\appendix

\section{Absorption probabilities in martingales}
\label{appA}

In the dynamical process described in the main text, $\hat\bs(t)$ determines $\bar\phi_a(\hat \bs(t))$ and thus determines $\hat\btheta_t$ and the distribution $p(\cdot|\hat\btheta_t)$. Consequently, given $\hat\bs(t)$ and thus  $\bar\phi_{a,t}\equiv \bar\phi_a(\hat \bs(t))$, we have
\begin{equation}
\label{eq:martingle }
\langle \bar\phi_{a,t+1}|\bar\phi_{a,t}\rangle_{\btheta_t}=\bar\phi_{a,t}\,.
\end{equation}
implying that $\bar\phi_{a,t}$ is a martingale. Since the sequence of samples $\hat \bs(0),\hat \bs(1),\cdots$ forms a Markov chain, so does $\bar\phi_{a,0}, \bar\phi_{a,1},\cdots$. 
Under the assumption that $\phi(s)$ is bounded, the states 
\begin{equation}
\label{eq:abs}
\phi_{a}^{+}=\max_{s}\phi_{a}(s),\qquad \phi_{a}^{-}=\min_{s}\phi_{a}(s)
\end{equation}
are absorbing states, for any Markov chain that is also a martingale. In other words, $P\{\bar\phi_{a,t+1}|\bar\phi_{a,t}=\phi_{a}^{\pm}\}=1$ if 
$\bar\phi_{a,t+1}=\phi_{a}^{\pm}$ and $P\{\bar\phi_{a,t+1}|\bar\phi_{a,t}=\phi_{a}^{\pm}\}=0$ otherwise. In addition, if there are no other absorbing states, the asymptotic probability to be absorbed at $\phi_{a}^{\pm}$ starting from $\bar\phi_{a,0}$ is given by
\begin{equation}
\label{absP}
\lim_{t\to\infty}P\{\bar\phi_{a,t}=\phi_a^+|\bar\phi_{a,0}\}=\frac{\bar\phi_{a,0}-\phi_{a}^{-}}{\phi_{a}^{+}-\phi_{a}^{-}}
\end{equation}
The proof of these statements is straightforward; we recall the argument in Feller~\cite{feller1968introduction} (p. 399). 
For a single variable $\phi$ taking discrete values in $[\phi_-,\phi_+]$, let $p_{\phi,\phi'}=P(\phi_{t+1}=\phi'|\phi_t=\phi)$ be the transition probability matrix. The martingale property implies 
\[
\langle\phi_{t+1}|\phi_t\rangle=\sum_{\phi'}p_{\phi_t,\phi'} \phi'=\phi_t\,.
\]
For $\phi_t=\phi_-$ this necessarily implies that $p_{\phi_-,\phi'}=0$ for all $\phi'>\phi_t$ and $p_{\phi_-,\phi_-}=1$, so $\phi_-$ is an absorbing state. The same is true for $\phi_t=\phi_+$. 

Let $p_{\phi,\phi'}^{(t)}$ be the $\phi,\phi'$ matrix element of the $t^{\rm th}$ power of the transition matrix. If there is no other absorbing state, then $p_{\phi,\phi'}^{(t)}\to 0$ for $t\to\infty$, for all $\phi'\neq \phi_\pm$. The iteration of the martingale property infinite times then implies that 
\[
\lim_{t\to\infty}\left[p_{\phi_0,\phi_-}^{(t)}\phi_-+p_{\phi_0,\phi_+}^{(t)}\phi_+\right]=\phi_0
\]
Eq.~(\ref{absP}) follows from the fact that $p_{\phi_0,\phi_-}^{(t)}=1-p_{\phi_0,\phi_+}^{(t)}$.\\

{\bf Absorption probabilities in Ising model.} In the case of the Ising model, $\bar \phi_1(\hat \bs) = M^{-1}\sum^{M,N}_{l,j=1} \hat \sigma^l_j$. From Eq. \eqref{eq:abs} above, we have $\phi^{\pm}_1=\pm N$ and thus using Eq.~\eqref{absP} above, we retrieve Eq.~\eqref{eq:PphiisingJ0} in the main text for the case of $\theta_2 = 0$. If, instead $\theta_2 \neq 0$, we can write
\begin{equation}
\label{eq:FCIMp}
f(s|\theta_1,\theta_2) = \frac{1}{Z'}\exp\left[\frac{\theta_2}{2N}\left(\phi_1(s)-\phi^*\right )^2\right],\qquad \phi^*=-N\frac{\theta_1}{\theta_2}
\end{equation}
with $Z'$ a normalization constant. 
Consequently, $\theta_1$ and $\theta_2$ are both coupled to the same order parameter $\phi_1=\sum_i\sigma_i$.
For a given $N$ and $\phi^* =N-2k$ \red{with $k=0,1,2,\cdots N$}, $f(s|\theta_1,\theta_2)$ in Eq.\ \eqref{eq:FCIMp} the distribution of $\phi_1$ becomes more and more concentrated around $\phi^*$, as $\theta_2\to -\infty$ (with $\theta_1=-\theta_2\phi^*/N\to\infty$ also diverging). This limit is realized as soon as, at some time step $t$, $\phi_1(\hat s^{l}(t))=\phi^*$ for all data points $\hat s^(l)(t)$, so that $\bar\phi_{1,t}=\phi^*$. For all $t'>t$ we will also have $\bar\phi_{1,t'}=\phi^*$ and thus $\phi^*$ is an absorbing state for $\bar\phi_{1,t}$. As a result, for $\theta_1$ and $\theta_2$ diverging with a finite ratio $\theta_1/\theta_2= -\phi^*/N$, the system thus has $N+1$ absorbing states, corresponding to the $N+1$ different values that $\phi^*$ can take.\\

{\bf Absorption probabilities in REM.} Similar to the Ising model discussed in the previous section, the Random Energy model (REM) is also defined over configurations of $N$ spins $s=(\sigma_1,\ldots,\sigma_N)$ with $\sigma_i=\pm 1$. It assigns a different parameter -- the energy $E(s)$ -- to each of the $2^n$ spin configuration, and the probability of each state is proportional to $\exp[-E(s)]$. In the notation of Eq.~\eqref{eq:expfam} of the main text, for each configuration $s$, we have a corresponding $\theta_s$ and $\phi_s$ and
\begin{equation}
\label{eq:REM}
\phi_{s}(s') = \delta_{s,s'}
\end{equation}
The model is an over-parametrized model in the sense that there is corresponding parameter for each spin configuration. 

Closed loop learning in this case is straightforward. For $\hat \bs$ comprised of $M$ data points, let us denote by $k_s$ the number of times configuration $s$ has been observed. Since $\sum_s k_s=M$, we have $\bar \phi_s(\hat \bs) = k_s$. 
Starting from a sample where state $s$ is observed $k_{s,0}$ times, with $\sum_s k_{s,0}=M$, at iteration $t$, let $k_{s,t}$ be the count of state $s$. Maximum likelihood returns the estimate $\exp[\hat\theta^t_{s}] /Z=\mu_t(s)=\frac{k_s,t}{M}$. From this a new sample $\hat s^{(t+1)}$ is drawn. It is easy to see that if a state $s$ fails to be observed at time $t$, i.e. if $k_{s,t}=0$, then $\hat\theta_{s,t}\to-\infty$, which means that state $s$ will never be observed forever after. Then the number of states with $k_{s,t}>0$ is a non-increasing function of time. The asymptotic state will be one where $k_{s^*,t}=M$ for one state and $k_{s,t}=0$ for all $s\neq s^*$. Alternatively, by noting that
for each $s$, $\bar \phi^{+}_s = M$ and $\bar \phi^{+}_s = 0$, and that $\bar \phi_{s,t} = k_{s,t}$, Eq.~(\ref{absP}) \red{above} can then be written as
\begin{equation}
\lim_{t\to\infty} P\{k_{s,t}=M| k_{s,0}\}=\frac{k_{s,0}}{M}\,.
\end{equation}
\red{which is Eq.~\eqref{eq:rem} of the main text.} Eq.~(\ref{absP}) clearly applies to the reduced process $\tilde s=1$ for state $s$ and $\tilde s=0$ for any other state $s'\neq s$. 

\section{The continuous time limit}
\label{appB}

Let us begin by deriving an expression for the transition matrix
\begin{equation}
p(\btheta_{t+1}=\btheta'|\btheta_t=\btheta)
=\sum_{\hat \bs}f(\hat \bs|\btheta)\delta\left(\btheta'-\hat\btheta(\hat \bs)\right).
\end{equation}
In order to do so, we first write, assuming $M\gg 1$,
\begin{eqnarray}
p(\btheta_{t+1}=\btheta'|\btheta_t=\btheta)
& = & \sum_{\hat \bs}\frac{f(\hat \bs|\btheta)}{f(\hat \bs|\hat\btheta)}f(\hat \bs|\hat\btheta)\delta\left(\btheta'-\hat\btheta\right)\\
 & \cong &  \sum_{\hat \bs}e^{-\frac{M}{2}(\btheta-\hat\btheta)\hat J(\btheta-\hat\btheta)}
f(\hat \bs|\hat\btheta)\delta\left(\btheta'-\hat\btheta\right)\\
 & = & e^{-\frac{M}{2}(\btheta-\btheta') J'(\btheta-\btheta')} \sum_{\hat \bs}
f(\hat \bs|\hat\btheta)\delta\left(\btheta'-\hat\btheta\right)\label{eq3}
\end{eqnarray}
here $\hat\btheta=\hat\btheta(\hat \bs)$ is a function of $\hat \bs$. In the first line, the term $\frac{f(\hat \bs|\btheta)}{f(\hat \bs|\hat\btheta)}$ is non-zero only for $\btheta\approx\hat \btheta$, so we expand the (log of the) numerator around $\hat\btheta$ and retain only terms up to 2nd order (with an error which is of order $1/M$). In the second line, 
\begin{equation}
\label{FIMapp}
J_{a,b}(\btheta)=\frac{\partial^2}{\partial\theta_a\partial\theta_b}\log Z=\langle \phi_a\phi_b\rangle_{\btheta}- \langle\phi_a\rangle_{\btheta}\langle\phi_b\rangle_{\btheta}
\end{equation}
is the Fisher Information matrix and we use the shorthands $\hat J=J(\hat\btheta)$ and $J'=J(\btheta')$. In Eq.~\eqref{eq3} above we substitute $\hat \btheta$ with $\btheta'$, exploiting the presence of the $\delta$-function. For the last term, we use the identity 
\begin{equation}
\int d\btheta g(\btheta)=\sum_{\hat\bs}\int d\btheta g(\btheta)f(\hat\bs|\btheta)
\cong \left(\frac{2\pi}{M}\right)^{D/2}\sum_{\hat{\bs}} \frac{f(\hat \bs|\hat\btheta)g(\hat\btheta)}{\sqrt{{\rm det} J(\hat\btheta)}}
\end{equation}
where $g(\btheta)$ is a generic function. This identity follows by estimating the integral in the middle expression by the saddle point method for $M\gg 1$. 
Using $g(\btheta)= \left(\frac{2\pi}{M}\right)^{-d/2}\sqrt{{\rm det} J(\btheta)}\,\delta(\btheta-\btheta')$ in Eq.~(\ref{eq3}), we find
\begin{eqnarray}
\label{wtrans0}
p(\btheta_{t+1}&=&\btheta'|\btheta_t=\btheta)\cong\left(\frac{M}{2\pi}\right)^{-D/2}\sqrt{{\rm det} J'}e^{-\frac{M}{2}(\btheta-\btheta')J'(\btheta-\btheta')}\\
&\cong & \left(\frac{M}{2\pi}\right)^{-D/2}\sqrt{{\rm det} J}e^{-\frac{M}{2}(\btheta'-\btheta)J(\btheta'-\btheta)-\frac{1}{2}(\btheta'-\btheta)\balpha}
\end{eqnarray}
where we expanded the term $\sqrt{{\rm det} J'}\cong \sqrt{{\rm det} J}\,e^{\frac{1}{2}(\btheta-\btheta')\balpha}$ with
\[
\alpha_a(\btheta)=\frac{\partial}{\partial\theta_a}\log {\rm det} J(\btheta)\,,
\]
and substituted $J'$ with $J$ in the quadratic term at the exponent, neglecting sub-leading corrections. This is a Gaussian distribution for $\btheta'$ with mean and variance
\begin{equation}
\langle\theta_{a}'\rangle_{\btheta}=\theta_a-\frac{1}{2M}\sum_bJ^{-1}_{a,b}\frac{\partial}{\partial\theta_b}\log {\rm det} J(\btheta)\,,\qquad \langle(\theta_{a}'-\langle\theta_{a}'\rangle_{\btheta})(\theta_b'-\langle\theta_{b}'\rangle_{\btheta})\rangle_{\btheta}=J^{-1}_{a,b}\,.
\end{equation}
In the continuous time limit, this leads to the stochastic differential equation
\begin{equation}
d\theta_a=-\frac{1}{2}\sum_bJ^{-1}_{a,b}\frac{\partial}{\partial\theta_b}\log {\rm det} J(\btheta)d\tau+dW_a\,,~~~\langle dW_{a,\tau}dW_{a,\tau'}\rangle_{\btheta_\tau}=J^{-1}_{a,b}(\btheta_\tau)\delta(\tau-\tau')d\tau\,.
\end{equation}
This corresponds to a  stochastic differential equation for the expected value of the sufficient statistics $\varphi_a\equiv\langle\phi_a(\bs)\rangle_{\btheta}=\frac{\partial\log Z}{\partial\theta_a}$ that is readily obtained with Ito calculus
\begin{eqnarray}
    d\varphi_a & = & \sum_b\frac{\partial\varphi_a}{\partial\theta_b}d\theta_b+\frac 1 2 \sum_{b,c} \langle dW_{b,\tau}dW_{c,\tau'}\rangle_{\btheta_\tau}\frac{\partial^2\varphi_a}{\partial\theta_b\partial\theta_c}d\tau\\
    &=&\sum_bJ_{a,b}\left[-\frac 1 2\sum_c J^{-1}_{b,c}\frac{\partial\log{\rm det}\hat J}{\partial\theta_c}d\tau+dW_b\right]+\frac 1 2 \sum_{b,c} J_{b,c}^{-1}\frac{\partial^2\varphi_a}{\partial\theta_b\partial\theta_c}d\tau\\
    &=&d\xi_{a,\tau}\label{eq:varphiSD}
\end{eqnarray}
where $d\xi_{a,\tau}=\sum_b J_{a,b}dW_{b,\tau}$ is a stochastic noise differential with covariance
\begin{equation}
    \langle d\xi_{a,\tau}'d\xi_{b,\tau'}'\rangle_{\bm{\varphi}_\tau}=J_{a,b}\delta(\tau-\tau')d\tau
\end{equation}
In the derivation of Eq.~\eqref{eq:varphiSD} we used the identities
\[
\frac{\partial}{\partial\theta_a}\log {\rm det} J(\btheta)=\sum_{b,c}J_{b,c}^{-1}\frac{\partial J_{b,c}}{\partial\theta_a}\,,\qquad\frac{\partial^2\varphi_a}{\partial\theta_b\partial\theta_c}=
\frac{\partial^3\log Z}{\partial\theta_a\partial\theta_b\partial\theta_c}=
\frac{\partial J_{b,c}}{\partial\theta_a}\,.
\]
Note that Eq.~\eqref{eq:varphiSD} is drift-less, as it should be given the martingale property. 

In the general case, the transition matrix is given by 
\begin{equation}
\label{eq:w}    
p(\btheta_{t+1}=\btheta'|\btheta_t=\btheta)=\sum_{\hbs,\hat\bt}p(\hat\bs|\btheta)q(\hat\bt)\delta\left(\btheta'-\bar\btheta(\hat\bs,\hat\bt)\right)
\end{equation}
where 
$\bar\btheta$ is determined via 
\begin{equation}
\label{eq:G}
    \bar\btheta(\hat\bs,\hat\bt)={\rm arg}\max_{\bvartheta} \mathcal{G}(\bvartheta,\hat\bs,\hat\bt)\,,\qquad \mathcal{G}(\bvartheta,\hat\bs,\hat\bt)=\log f(\hat\bs|\bvartheta)+\log f(\hat\bt|\bvartheta)+u\log p_0(\bvartheta)
\end{equation}
With $m=u=0$ this problem reduces to the case where $\bar\btheta=\hat\btheta(\hat\bs)$ is the maximum likelihood (ML) estimator and all samples are generated from the ML distribution at the previous step. With $m>0$ this problem envisages the possibility that some samples $\hat\bt$ are generated from the ``true'' distribution $q(\cdot)$ while the others are generated from $\bar\btheta$ and with $u=1$ we look at the case where $\bar\btheta$ is the maximum {\em a posteriori} (MAP) estimator, with prior $p_0$.

With finite $m$ and $M\gg 1$, the maximum is dominated by the ML estimator $\hat\btheta(\hat\bs)$, so we expand around it:
\begin{equation}
    \mathcal{G}(\bvartheta)\simeq \log \left[f(\hat\bs|\hat\btheta)f(\hat\bt|\hat\btheta)\right]+u\log p_0(\hat\btheta)
    -\frac M 2 (\bvartheta-\hat\btheta)^T \hat \bJ(\bvartheta-\hat\btheta) + \balpha^T (\bvartheta-\hat\btheta)+\ldots\,,
\end{equation}
where $\hat \bJ=\bJ(\hat\btheta)$ and $\tilde\balpha\in\mathbb{R}^d$ is a vector with components 
\begin{equation}
    \tilde\alpha_a = \left.\frac{\partial}{\partial\vartheta_a}\log f(\hat\bt|\bvartheta)\right|_{\hat\btheta}+u\left.\frac{\partial}{\partial\vartheta_a}\log p_0(\bvartheta)\right|_{\hat\btheta}\,.
\end{equation}
To leading order in $M$, the maximum is given by
\begin{equation}
\label{eq:bartheta}
    \bar\btheta(\bs,\bt)=\hbtheta(\bs)+\frac{\hat \bJ^{-1}\tilde\balpha}{M}\,.
\end{equation}
Therefore, in the continuous time limit, the stochastic differential equation acquires an additional drift
\[
{\bJ^{-1}\langle\tilde\balpha\rangle_{\btheta}}=\bJ^{-1}\frac{\partial}{\partial\btheta}\left[u\log p_0(\btheta)-m\, {\rm D_{KL}}(q|\btheta)\right]
\], 
so that
\begin{eqnarray}
    d\theta_a&=&-\sum_b J^{-1}_{a,b}\frac{\partial}{\partial\theta_b}\left[\frac{1}{2}\log {\rm det} J(\btheta)-u\log p_0(\btheta)+mD(q|\btheta)\right]d\tau+dW_a\,,\\
    &~&\langle dW_{a,\tau}dW_{a,\tau'}\rangle_{\btheta}=J^{-1}_{a,b}(\btheta)\delta(\tau-\tau')\, d\tau
\end{eqnarray}
as in Eqs.~\eqref{eq:dtheta}-\eqref{eq:dW} of the main text.

\subsection{The stationary state}

As discussed in Gardiner~\cite{gardiner2009stochastic}, the Fokker-Planck equation associated to the stochastic differential equation \eqref{eq:dtheta} in the Letter can be written as
\begin{equation}
    \frac{\partial p}{\partial t}=\sum_a\frac{\partial F_a}{\partial\theta_a}
\end{equation}
where the probability current $\mathbf{F}$ is given by
\begin{equation}
    F_a=-A_a p+\frac 1 2 \sum_b\frac{\partial}{\partial\theta_b}J^{-1}_{a,b}p=-\left[A_a -\frac 1 2 \sum_b\frac{\partial J^{-1}_{a,b}}{\partial\theta_b}\right]p+\frac 1 2 \sum_b J^{-1}_{a,b}\frac{\partial p}{\partial\theta_b}
\end{equation}
The zero current condition $\mathbf{F}=0$ reads
\begin{equation}
    2A_a -\sum_b\frac{\partial J^{-1}_{a,b}}{\partial\theta_b}=\sum_b J^{-1}_{a,b}\frac{\partial \log p}{\partial\theta_b}
\end{equation}
or
\begin{eqnarray}
    \frac{\partial \log p}{\partial\theta_b} &=& \sum_a J_{b,a}\left[2A_a -\sum_c\frac{\partial J^{-1}_{a,c}}{\partial\theta_c}\right]\\
    & = & -2\frac{\partial U}{\partial\theta_b}-\sum_{a,c} J_{b,a}\frac{\partial J^{-1}_{a,c}}{\partial\theta_c}\\
    & = & -2\frac{\partial U}{\partial\theta_b}+\sum_{a,c} J^{-1}_{a,c}\frac{\partial J_{a,b}}{\partial\theta_c}\\
    & = & -2\frac{\partial U}{\partial\theta_b}+\sum_{a,c} \frac{\partial\log{\rm det} J}{\partial J_{a,c}}\frac{\partial J_{a,c}}{\partial\theta_b}\\
    & = & \frac{\partial }{\partial\theta_b}\left[-2U+\log{\rm det}J\right]
\end{eqnarray}
where 
\[
U(\btheta) = \frac 1 2 \log{\rm det}J-u\log p_0(\btheta)+m\, {\rm D_{KL}}(q|\btheta)
\]
is the same as Eq.~\eqref{eq:U} \red{in the main text} and we used the identities
\begin{eqnarray}
\label{eq:idJ2}
    \sum_{a,c} J_{b,a}\frac{\partial J^{-1}_{a,c}}{\partial\theta_c} &=& \frac{\partial}{\partial\theta_c}\left[\sum_{a,c} J_{b,a}J^{-1}_{a,c}\right]-\sum_{a,c} \frac{\partial J_{a,b}}{\partial\theta_c}J^{-1}_{a,c}=-\sum_{a,c} J^{-1}_{a,c}\frac{\partial J_{a,b}}{\partial\theta_c}\\
    \frac{\partial J_{a,b}}{\partial\theta_c}&=&\frac{\partial^3 \log Z}{\partial\theta_a\partial\theta_b\partial\theta_c}=\frac{\partial J_{a,c}}{\partial\theta_b}
\label{eq:idJ1}   \\ 
    J^{-1}_{a,c} &=& \frac{\partial\log{\rm det} J}{\partial J_{a,c}}
\end{eqnarray}
where $Z$ is the normalization constant in Eq.~\eqref{eq:expfam} of the main text. In Eq.~(\ref{eq:idJ2}) we used the fact that the term in square brackets is the identity matrix and hence its derivative vanishes, and then we used identity~(\ref{eq:idJ1}) to swap indices $c\leftrightarrow b$.

Therefore, the stationary state distribution is
\begin{equation}
    \label{eq:pss}
    p_{st}(\btheta)= c\,{\rm det}J\,e^{-2U(\btheta)}
\end{equation}
with $c$ a normalization constant. The stationary state exists as long as $p_{st}$ is normalizable, i.e. as long as $c$ is finite. This yields Eq.~\eqref{eq:pstat} in the main text.

\begin{figure}
    \centering
    \includegraphics[width=0.6\linewidth]{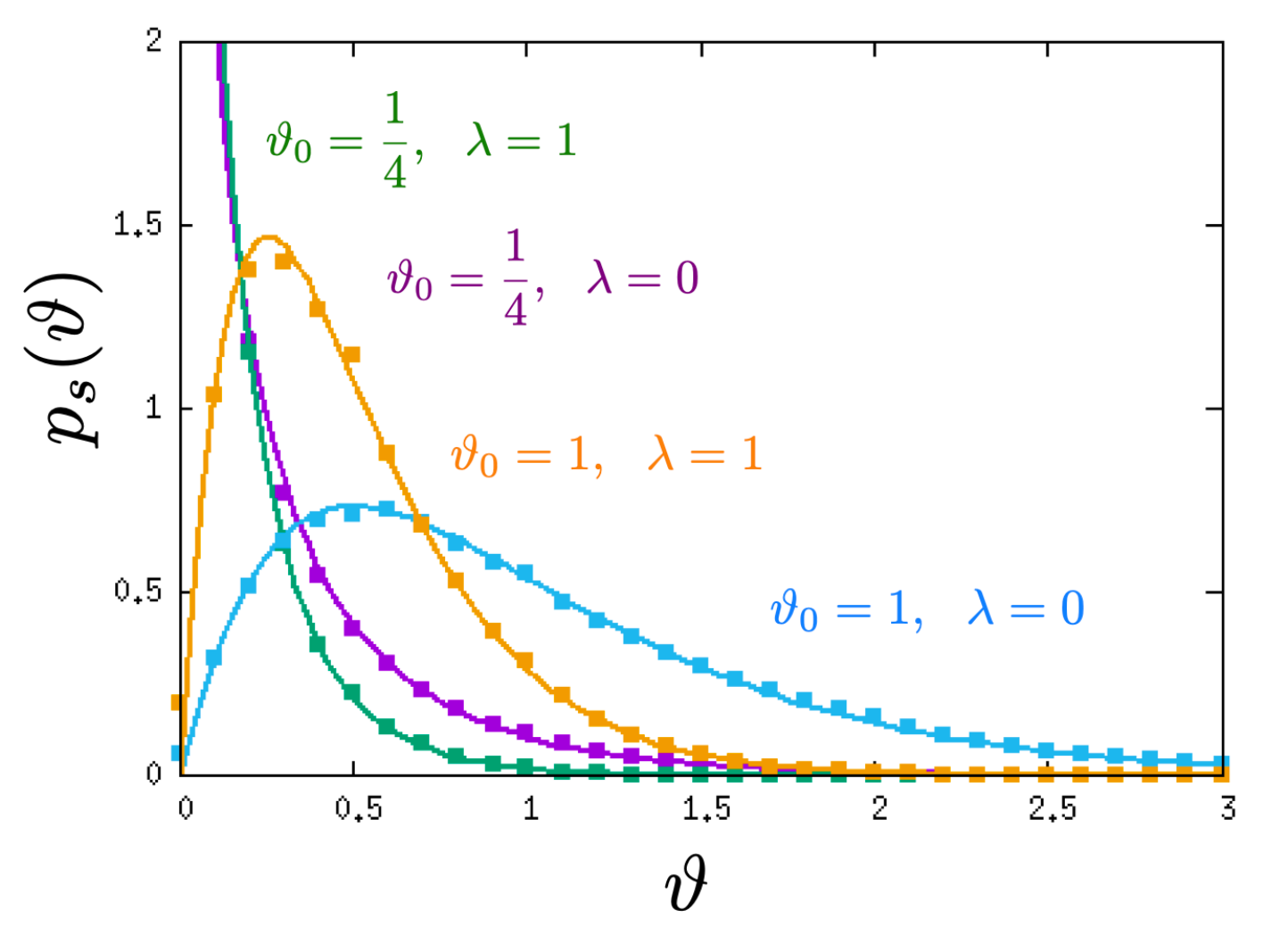}
    \caption{\red{Distribution of $\vartheta$ for the Poisson distribution with $m=1$, $M=200$ and different values of $\lambda$ controlling the prior, and $\vartheta_0$ the mean of the external data from the distribution $q$. Points refer to sampling of the closed-loop retraining process and full lines refer to the analytical solution Eq.~(\ref{eq_pspoiapp}). As discussed in the text below, when $\lambda \to 0$, we have the case of $u=0$ that is no regularization or priors, and thus with $m=0$, model collapse occurs. The results in this figure shows how adding one data point prevents this.}}
    \label{fig:m}
\end{figure}

\section{The case of Poisson distribution}
\label{appC}

Let us consider the case of a Poisson distribution
\[
p(s|\vartheta)=\frac{\vartheta^{s}}{s !}e^{-\vartheta}
\]
with mean $\langle\bs\rangle_\vartheta=\vartheta\ge 0$ and prior $p_0(\vartheta)=\lambda e^{-\lambda\vartheta}$. Since the distribution of $s$ depends on a single parameter we do not use boldface characters for parameters. We use notation $\vartheta$ instead of $\theta$ because $p(s|\vartheta)$ is an EF but it is not expressed in terms of the natural parameter $\theta=\log\vartheta$, which is also why we do not use the notation $f(s|\theta)$. Finally, to simplify the notation, we shall redefine $u\lambda=\lambda$, so that the case $u=0$ is obtained in the limit $\lambda\to 0$.

With $p_0(\vartheta)=\lambda e^{-\lambda\vartheta}$ and $\hat\bt$ being $m$ i.i.d. samples from a Poisson distribution with mean $\vartheta_0$, the optimization of 
\[
\mathcal{G}(\bvartheta,\hat\bs,\hat\bt)=\log p(\hat\bs|\bvartheta)+\log p(\hat\bt|\bvartheta)+\log p_0(\bvartheta)
\]
yields
\[
\frac{\partial\mathcal{G}}{\partial\vartheta}=\frac{S+T}{\vartheta}-M-m-\lambda~~~\Rightarrow~~~ \bar\vartheta=\frac{S+T}{M+m+\lambda},\qquad S=\sum_{i=1}^M \hat{s}_i,~~T=\sum_{i=1}^m \hat{t}_i\,.
\]
Notice that $\bar\vartheta=0$ is not an absorbing state when $m>0$ because even if $S=T=0$ at some $t$, so that $\vartheta_{t+1}=0$, the $m$ samples $\hat\bt$ will again be drawn from a Poisson distribution with mean $\vartheta_0>0$. Conversely, for $m=0$ as soon as $S=0$ for some $t$, $\vartheta_{t'}=0$ for all $t'>t$.

The conditional distribution of $\vartheta'$ given $\vartheta$ is given by
\begin{eqnarray*}
p(\vartheta_{t+1}=\vartheta'|\vartheta_t=\vartheta) &= & \sum_{S=0}^\infty \frac{(M\vartheta)^S}{S!}e^{-M\vartheta}\sum_{T=0}^\infty \frac{(m\vartheta_0)^S}{S!}e^{-M\vartheta} \delta\left(\vartheta'-\frac{S+T}{M+n+\lambda}\right)\\
 & = & \int_{-\infty}^\infty\frac{dk}{2\pi}e^{ik\vartheta'} \exp\left[M\vartheta\left(1-e^{-ik/(M+m+\lambda)}\right)\right] \exp\left[m\vartheta_0\left(1-e^{-ik/(M+m+\lambda)}\right)\right]\\
 & \cong & \int_{-\infty}^\infty\frac{dk}{2\pi}\exp\left[ik\left(\vartheta'-\frac{M\vartheta+m\vartheta_0}{M+m+\lambda}\right)-\frac{M\vartheta+m\vartheta_0}{2(M+m+\lambda)^2}k^2+O(k^3/M^2)\right]\\
 & \cong & \frac{M+m+\lambda}{\sqrt{2\pi(M\vartheta+m\vartheta_0)}}\exp\left[-\frac 1 2 \frac{(\vartheta'-\langle\vartheta'\rangle_\vartheta)^2}{\langle (\vartheta'-\langle\vartheta'\rangle_\vartheta)^2\rangle}\right]
\end{eqnarray*}
which is a Gaussian with mean and variance
\[
\langle\vartheta'\rangle_\vartheta=\frac{M\vartheta+m\vartheta_0}{M+m+\lambda}=\vartheta+\frac{m+\lambda}{M+m+\lambda}\left(\frac{m\vartheta_0}{m+\lambda}-\vartheta\right)\,,\qquad 
\langle (\vartheta'-\langle\vartheta'\rangle_\vartheta)^2\rangle=\frac{M\vartheta+m\vartheta_0}{(M+m+\lambda)^2}
\]
respectively. In the limit $M\to\infty$, in rescaled time, the parameter $\vartheta$ follows the stochastic differential equation 
\begin{equation}
d\vartheta_\tau=(m+\lambda)\left[\frac{m\vartheta_0}{m+\lambda}-\vartheta_\tau\right]d\tau+\sqrt{\vartheta_\tau}dW
\end{equation}
which is a Cox-Ingersoll-Ross model~\cite{cox1985theory}. When $2m\vartheta_0> 0$ the process admits a stationary distribution
\begin{equation}
\label{eq_pspoiapp}
p_{st}(\vartheta)=\frac{[2(m+\lambda)]^{2m\vartheta_0}}{\Gamma(2m\vartheta_0)}\vartheta^{2m\vartheta_0-1}e^{-2(m+\lambda)\vartheta}
\end{equation}
which is given by a Gamma distribution.

The results of Eq.~\eqref{eq_pspoiapp} are in excellent agreement with numerical simulations, as shown in Fig.~\ref{fig:m}. Note that with $m=0$ the distribution is not normalizable, i.e. $\vartheta=0$ is an absorbing state for all $\lambda>0$ when $m=0$. Finally we note that when the Poisson distribution is expressed in terms of the canonical parameter $\theta=\log\langle s\rangle_{\theta}$, Eq.~(\ref{eq:pss}) leads to a stationary state
\begin{equation}
    p_{st}(\theta) = \frac{[2(m+\lambda)]^{2m\vartheta_0}}{\Gamma(2m\vartheta_0)}e^{2m\vartheta_0\theta-2(m+\lambda)e^{\theta}}
\end{equation}
that coincides with Eq.~(\ref{eq:pstat}) of the main text, because ${\rm D_{KL}}(q|\theta)=\vartheta_0\log\frac{\vartheta_0}{\vartheta}+\vartheta-\vartheta_0$ in this case.

\end{document}